\documentclass[a4paper,fleqn]{cas-dc}

\usepackage[numbers]{natbib}

\usepackage{soul} 
\usepackage{algorithm}
\usepackage{algpseudocode}

\def\tsc#1{\csdef{#1}{\textsc{\lowercase{#1}}\xspace}}
\tsc{WGM}
\tsc{QE}
\tsc{EP}
\tsc{PMS}
\tsc{BEC}
\tsc{DE}

\begin{document}
\let\WriteBookmarks\relax
\def\floatpagepagefraction{1}
\def\textpagefraction{.001}

\shorttitle{Fast Humanoid Robot Door Traversals}

\shortauthors{Calvert et~al.}

\title [mode = title]{A Behavior Architecture for Fast Humanoid Robot Door Traversals}                      
\tnotemark[1]

\tnotetext[1]{This work was funded through ONR Grant 2023-004-01.}

\author[1,2]{Duncan Calvert}[orcid=0009-0001-5255-6882]
\ead{dcalvert@ihmc.org} 

\author[1]{Luigi Penco}[orcid=0000-0002-2938-2546]
\ead{lpenco@ihmc.org}

\author[1, 2]{Dexton Anderson}[orcid=0009-0004-1402-0536]
\ead{danderson@ihmc.org}

\author[1, 2]{Tomasz Bialek}[orcid=0009-0008-1411-2376]
\ead{tbialek@ihmc.org}

\author[1, 2]{Arghya Chatterjee}[orcid=]
\ead{achatterjee@ihmc.org}

\author[1, 2]{Bhavyansh Mishra}[orcid=]
\ead{bmishra@ihmc.org}

\author[1, 2]{Geoffrey Clark}[orcid=]
\ead{gclark@ihmc.org}

\author[1]{Sylvain Bertrand}[orcid=]
\ead{sbertrand@ihmc.org}

\author[1, 2]{Robert Griffin}[orcid=]
\ead{rgriffin@ihmc.org}

\affiliation[1]{organization={Florida Institute for Human and Machine Cognition},
    addressline={40 South Alcaniz St.}, 
    city={Pensacola, FL},
    postcode={32502}, 
    state={FL},
    country={USA}}

\affiliation[2]{organization={University of West Florida},
    addressline={11000 University Pkwy}, 
    city={Pensacola, FL},
    postcode={32514}, 
    state={FL},
    country={USA}}

\begin{abstract}
Towards the role of humanoid robots as squad mates in urban operations and other domains, we identified doors as a major area lacking capability development.
In this paper, we focus on the ability of humanoid robots to navigate and deal with doors.
Human-sized doors are ubiquitous in many environment domains and the humanoid form factor is uniquely suited to operate and traverse them.
We present an architecture which incorporates GPU accelerated perception and a tree based interactive behavior coordination system with a whole body motion and walking controller.
Our system is capable of performing door traversals on a variety of door types.
It supports rapid authoring of behaviors for unseen door types and techniques to achieve re-usability of those authored behaviors.
The behaviors are modelled using trees and feature logical reactivity and action sequences that can be executed with layered concurrency to increase speed.
Primitive actions are built on top of our existing whole body controller which supports manipulation while walking.
We include a perception system using both neural networks and classical computer vision for door mechanism detection outside of the lab environment.
We present operator-robot interdependence analysis charts to explore how human cognition is combined with artificial intelligence to produce complex robot behavior.
Finally, we present and discuss real robot performances of fast door traversals on our Nadia humanoid robot.
Videos online at https://www.youtube.com/playlist?list=PLXuyT8w3JVgMPaB5nWNRNHtqzRK8i68dy.
\end{abstract}




\maketitle

\section{Introduction}

The humanoid form has uniquely diverse mobility and manipulation capabilities that drive its suitability as the embodiment of a general purpose robot.
This has lead to the pursuit of building humanoid robots to perform useful tasks in spaces designed for humans.
One component that is lacking in the literature is how robots can quickly and effectively perform useful whole-body loco-manipulation tasks automatically.
There are many challenges in building a system for generating and executing fast loco-manipulation behaviors.
Part of the challenge is being able to create behaviors for a wide variety of tasks and to do so in a short timeline as needed.
We present an authoring system which contains many features that when put together, provide sufficient capability in this regard.
Another aspect of the challenge is in getting the behaviors working on a real robot.
We also present an integrated system for testing and iterating on the execution of these behaviors on the real robot.

This topic of behavioral system integration is important to the understanding of how to build useful humanoid robotics systems.
There are many aspects of design and some components, if overlooked or misunderstood, could have a negative impact on performance or long term results.
As society builds more humanoid robots for military and industrial purposes, this research could help in grounding system level design principles.
Furthermore, we believe this topic of research does not lose its relevance when modern AI approaches are taken, as we think many of the core functionalities and principles explored in this paper are fundamental to humanoid robot loco-manipulation.

A key element to this is \textit{behavior generation} -- assembling motor skills to compose tasks and assembling tasks to compose behaviors.
In this work we utilize an operator to author behaviors as a tree structure of nodes with general capabilities.
This paper explores techniques that build upon known, useful principles in the literature in an effort to nudge the state of the art of behavior authoring on humanoid robots forward.
We place a core focus on speeding up the authoring process of behaviors by supporting online behavior generation and modification.
By providing a framework for building and modifying behavior trees and action sequences at runtime, most behavioral changes can be done online and quickly enacted on the robot.
The increase in authoring speed has allowed us to create more than twenty varieties of working loco-manipulation behaviors on the real robot in the span of one year.
The behaviors are primarily different types of door traversals but debris clearing behaviors such as moving trash cans and couches are also included.

Our behavior coordination architecture is designed to create fast, resilient, and reusable loco-manpipulation behaviors for humanoid robots as part of a larger effort towards robot that do useful and economical work in the real world.
To achieve this, we incorporated state-of-the-art methodologies in building a uniquely complete set of functionalities.
The concepts of affordance templates, action primitives, sequential composition, concurrent action layering, behavior trees, and human-robot interdependence were carefully included into the presented novel implementation.

This is currently no body of research available that explores door traversals with bipedal humanoid robots.
There are many publications which explore door traversals on wheeled base robots with arms, such as \cite{Jain2008dooropening}, \cite{2010ChittaDoorOpening} \cite{xiong2024adaptive}, and \cite{kang2024door}.
We believe this is a critical area of study, as the humanoid form factor is especially well-suited for navigating complex environments encountered in disaster response, space exploration, and urban settings.
The advantages of the humanoid structure over wheeled and other mobile platforms are explored further in \cite{Johnson_2017}.
Additionally, in this work, we find there are important differences in the planning required for wheeled base and legged robots.

\begin{figure*}[ht]
    \centering
        \includegraphics[width=2.0\columnwidth]{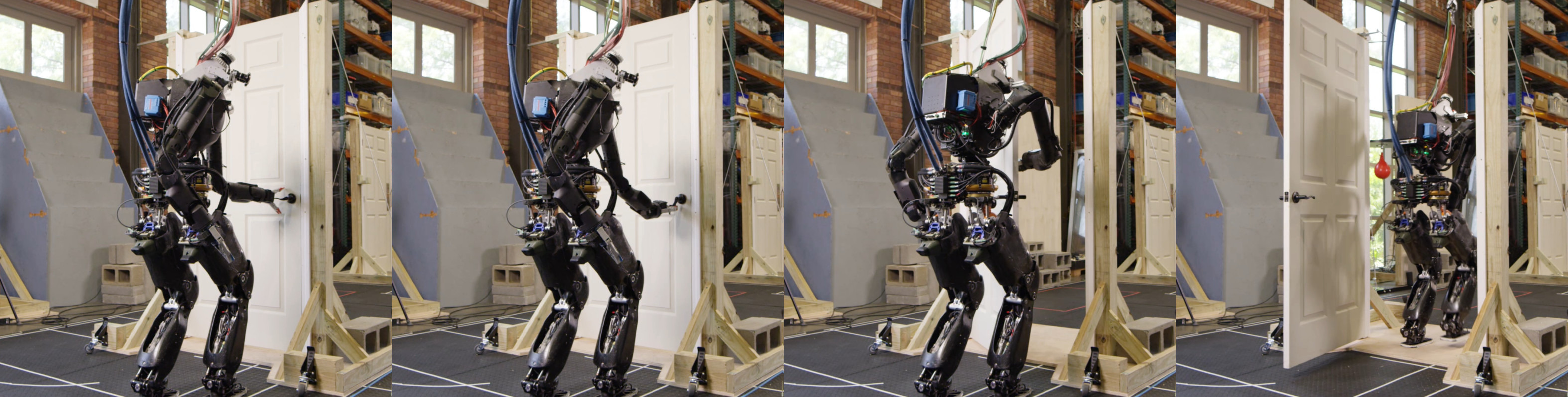}
    \caption{The Nadia humanoid robot performing a right pull lever handle door traversal using cycloidal drive forearms and Sake grippers.
    Corresponding video: 2024NadiaRightPullLeverHandleCycloidLowRes.mp4}
    \label{fig:2024NadiaRightPullLeverHandleCycloid}
\end{figure*}

To build modular and adaptable behaviors, we structured them as a composition of low-level actions.
May of these low-level actions are derived from the Affordance Template (AT) framework\cite{Hart_2014, Hart_2015}.
ATs are generally a sequence of primitive tasks that are parameterized to a type of higher level task, referred to as an affordance.
For example, an affordance template for opening a drawer could be a set of grasp approach, grasp, pull drawer, and let go actions that are available to execute any time the robot see a certain type of drawer.
It is a way of defining general action plans for ahead of time for high level tasks the robot is expected to encounter.
Additionally, the AT framework provides an integrated environment for authoring templates and provides a general definition language for robot-agnostic manipulation.

The AT framework is also naturally extensible.
In recent years, advanced planners such as for stance and grasp generation, navigation, and motion planning have been integrated\cite{Hart_2022}.
There is also a growing area termed ``affordance primitives''\cite{Pettinger_2020, Pettinger_2022} (APs) which are a way to model actions with constrained movement, such as turning a valve or closing a drawer.
A special AP called a screw primitive has been used to define impedance controlled manipulation of valves and drawers.

ATs have been used on NASA's Valykrie humanoid robot to perform a car door opening, retrieve an improvised explosive device (IED) bag, drop the IED bag off in a total containment vessel (TCV) tray, and push the TCV tray in and push it's button\cite{2019_Jorgensen_valEOD}.
The operator could load, edit, and execute ATs on the fly which provided a human-in-the-loop environment.

To coordinate the ATs, and more in general the low-level actions, we employ Behavior Trees\cite{2018_colledanchise, 2022_iovino_behavior_trees} (BTs), which are a popular state-of-the-art model for representing and orchestrating complex behavior.
BTs define logical operator nodes such as sequence, fallback, and parallel which work to control the execution flow of a behavior.
In BTs, actions are performed by the leaf nodes which command the robot and gather environmental data.
Finally, BTs provide reactivity though a ticking system in which each tick starts at the root node.
This is in contrast to state machines in which each tick starts from the previous state.
By continuously re-evaluating from the top down, from tick to tick there are pathways to ending up in very different parts of the tree without explicit connections between those parts.

Recently, BTs have been used to autonomously navigate cluttered terrain with a centaur wheeled-legged robot\cite{2023_centaur_bt}.
It demonstrates successful navigation on a real hardware system with legs.
The behavior tree in this work manages a two-level planner for coordinating intricate locomotion over a debris field of cinder blocks.
It includes reactivity to contacts and failures and decides on the next appropriate action to achieve a level of resiliency.
However, we are not aware of any prior work that illustrates an application of BTs to loco-manipulation behaviors on bipedal humanoid robots.

Behavior trees and low-level actions are integrated into an interface grounded in Coactive Design (CD)\cite{Johnson_2014, Johnson_2018} principles.
CD is a methodology which gives structure to the design and engineering of systems from the perspective of the interdependence of humans and machines.
CD is an iterative process that is comprised of three main steps: identifying, documenting, and iterating on interdependence relationships.
Interdependence analysis (IA) charts are used to document the interdependence relationships, which illustrate the connections between system components.
CD was previously used for the design and development of the operator interface used by IHMC in the 2015 DARPA Robotics Challenge (DRC) Finals.
A later analysis details how the methodology led to success in the competition\cite{Johnson_2017}.

MIT's DRC team developed an alternative system called Director\cite{Marion_2017} which used AT concepts in a framework to integrate robot autonomy components and task execution.
Director had increased autonomy compared to tools used by other teams in the DRC.
It featured advanced planners for manipulation and locomotion, an ``operator in the loop'' pipeline for executing actions, a 3D scene with interactive widgets, and a way to make custom panels of widgets.
A Python program editor was embedded directly in the user interface to write scripts for tasks at the behavior level.

\begin{figure*}[ht]
    \centering
        \includegraphics[width=2.0\columnwidth]{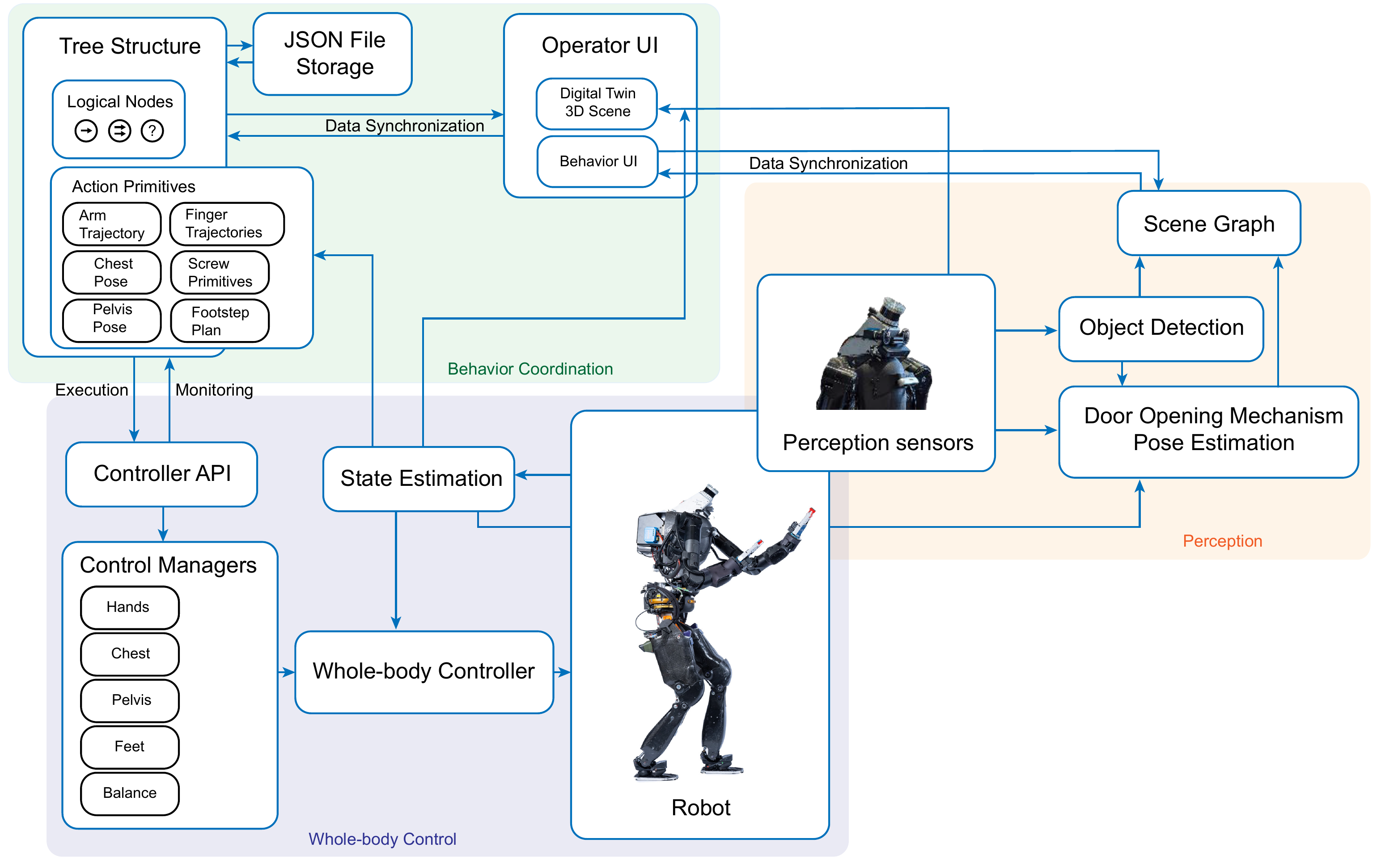}
    \caption{An all inclusive overview of the parts involved in this work.}
    \label{fig:Behavior_Overview}
\end{figure*}

A well-designed interface that supports the robot's directability and observability is crucial, but equally essential is the capability to accurately detect objects, as they often serve as key points of interaction during tasks.
YOLO-based methods have recently become popular for detecting doors\cite{bhatti2023yolodoor}.
When combined with the planar region extraction of the door panel, the position and orientation of door handles can be obtained.
\cite{kang2024door} uses YOLACT image segmentation to obtain the door handle position and a RANSAC plane segmentation algorithm to obtain the orientation.
\cite{Mishra_2021, mishra2024terrainmap} present a planar region extraction algorithm which can be used to detect door panels.
Together, these detection and segmentation methods provide the foundational perception capabilities needed for effective interaction with doors.

In this paper, we present a system for authoring and operating bipedal loco-manipulation behaviors.
To validate our approach, we present and demonstrate a variety of fast door traversal behaviors on real hardware.
Our implementation builds on state-of-the-art techniques and includes several novel contributions:

\begin{itemize}
    \item To enable increasing the speed of humanoid door traversal behaviors, we developed a system for an operator to compose concurrent sequences of locomotion and manipulation action primitives.
    \item Our integrated behavior framework enables the operator and robot to author reusable, fast, and resilient whole body loco-manipulation behaviors for types of doors that have not yet been encountered.
    \item To detect door opening hardware well enough to construct and perform door traversal behaviors outside of a lab environment, we developed a novel detection system that combines YOLO-based object detection and segmentation, depth correspondence, and rapid planar region extraction.
\end{itemize}

\section{Architecture Overview}

Our system, illustrated in \autoref{fig:Behavior_Overview}, spans three major domains: behavior coordination, perception, and whole body control.
We coordinate robot behavior using an interactive tree-based runtime which contains logical nodes and primitive actions.
The on-robot runtime is synchronized to an operator user interface containing a digital twin of the robot and environment, direct teleoperation elements, and a behavior management interface.
The behavior management interface facilitates managing both the authoring and execution of behaviors.
The perception system is a novel assembly of a YOLO neural network, planar region extractor, and heuristics for detecting door opening mechanisms.
These system elements interface with a model-based whole-body controller, walking algorithm, and the Nadia hardware platform.

\subsection{Tree-Based Runtime with Concurrency}

The tree structure is the base of the behavior representation, which is comprised of logical and action primitive nodes.
At runtime, logical evaluation starts at the root node and propagates downward as determined by the node implementations.
This logic is used to program reactivity and semantic decision making.
These nodes have access to perceived object frames so they are able to query knowledge about the environment and make decisions about what to do next.
When tasks fail, sub-sequences can be re-executed or alternative sub-trees of behavior can be invoked.

In contrast to BTs, our implementation updates all nodes on each tick and does not use predefined return types ("success", "failure", "running", etc).
This supports a more free-form implementation with less formal structure.
Instead, ours nodes implement a common interface for storing children, saving and loading with JSON files, executing on the robot side, and rendering user interface elements on the operator side.

\begin{figure}
    \centering
        \includegraphics[width=1.0\columnwidth]{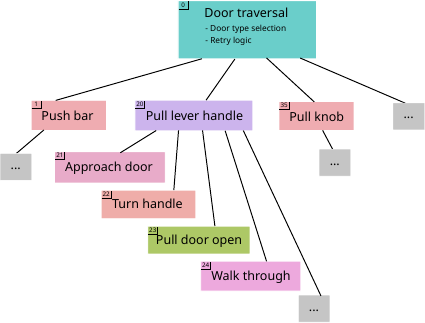}
    \caption{The general structure of a tree for door traversals. A high level node triages control to subtrees for particular door types and takes action in failure scenarios to retry set of actions or change strategy.}
    \label{fig:TreeStructure}
\end{figure}

\autoref{fig:TreeStructure} illustrates the tree structure used for door traversals.
At the root, we place a logical door traversal node which manages the various supported types door traversals.
It is able to trigger and monitor the execution of its children and access the scene knowledge graph from the perception system.
Reactive and robust logic is then programmed into the door traversal node which enables resiliency through retrying and employing alternative strategies in response to failures.

\begin{figure}
    \centering
        \includegraphics[width=1.0\columnwidth]{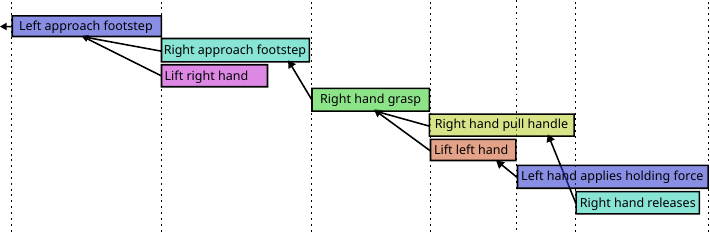}
    \caption{To achieve layered, concurrent execution while maintaining a sequence structure, each node has a pointer to a node to execute after. When an action is next for execution, it can start when the action it's waiting for is complete. The first part of a left pull handle door is illustrated as an example.}
    \label{fig:ActionLayeringIllustration}
\end{figure}

To speed up execution and enable more capability, our action executor features layered concurrency.
This is accomplished with an ``execute after'' field on each action primitive node which points to a prior action as a dependency on its completion.
This keeps actions starting in order but potentially earlier.
We illustrate this in \autoref{fig:ActionLayeringIllustration} and define Algorithm \autoref{alg:autonomous_execution}.

\begin{algorithm}
\caption{Autonomous Concurrent Layered Execution}
\label{alg:autonomous_execution}
\begin{algorithmic}
\State \textbf{Let} $A$ \textbf{be the set of actions} $\{a_0, a_1, a_2, \ldots\}$
\State $i \gets$ \text{index of the next action to execute}

\Function{update}{}
  \While{$i < \text{length}(A)$}
    \If{\Call{shouldExecute}{$a_i$}}
      \State \Call{executeAsynchronously}{$a_i$}
      \State $i \gets i + 1$
    \EndIf
  \EndWhile
\EndFunction

\Function{shouldExecute}{action}
  \State $a_{\text{after}} \gets$ \Call{getExecuteAfter}{action}
  \State \Return $\neg$ \Call{isExecuting}{$a_{\text{after}}$}
\EndFunction

\end{algorithmic}
\end{algorithm}

\subsection{Action Primitives}\label{section:action_primitives}

The primary interface between the behaviors and the controller are the action primitives, which are the leaf nodes in the tree and the pieces that trigger execution and handle monitoring execution.
The action primitives include footsteps and arm, hand, chest, and pelvis trajectories.
Through the authoring process, this primitives can be assembled into ATs.
The behavior author is responsible for assembling sequences of actions the ensure inherent stability when executed.

\begin{figure}
    \centering
        \includegraphics[width=0.7\columnwidth]{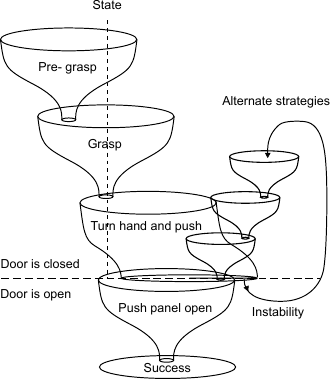}
    \caption{An illustration of how we apply sequential composition. During authoring, actions are carefully placed in an order in which the Cartesian result of the action informs the robot about the state of the environment such that it can automatically decide to proceed or not. When control is insufficient, the state will not fall into the stable region of the next action and a high level node will either retry or employ alternative strategies.}
    \label{fig:SequentialCompositionIllustration}
\end{figure}

\cite{1999_burridge} discusses the guarantees of stability of non-linear dynamic systems when multiple controllers are strung together.
It is described and illustrated as cascading Lyupanov funnels as shown in \autoref{fig:SequentialCompositionIllustration} where system states are accepted in boundaries at the top and are controlled to a state that is acceptable by the next controller in the sequence.
Towards modelling our system in this way, our primitive actions each implement entry and exit conditions.
On-entry conditions serve to gate action execution with checks that determine the feasibility while on-exit conditions determine success or failure.

\begin{figure}
    \centering
        \includegraphics[width=1.0\columnwidth]{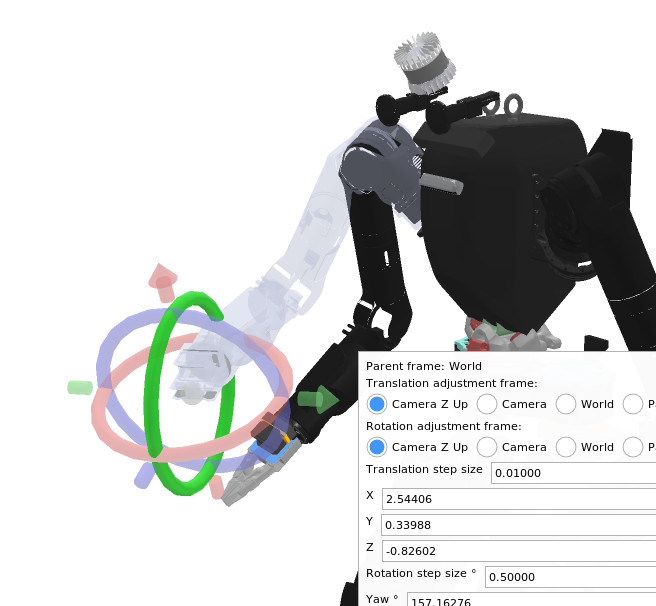}
    \caption{A gizmo for manipulating the hand's 6 DoF pose with a mouse and keyboard.
    The mouse can be used to drag the rings and arrows to perform constrained adjustments to those axes.
    The keyboard arrow keys can be used to move the gizmo with respect to the adjustment frames selected in the context menu.
    Modifier keys Ctrl, Alt, and Shift are used to select rotation axes and adjust the sensitivity.
    The context menu shown on the right allows the user to change the frame of adjustment, nudge the pose by small amounts, and view and edit the absolute values of the pose. The IK preview of the arm is also shown as a transparent graphic.}
    \label{fig:gizmo_composite}
\end{figure}

Furthermore, these actions are setup in task relative frames to provide re-usability in a particular environmental context.
We provide the behavior author with a 3D pose gizmo shown in \autoref{fig:gizmo_composite} to allow them to interactively edit the task relative transforms.

To orchestrate walking, we implemented a footstep plan action that represents a sequence of footsteps that get queued to the controller.
Frame-relative footsteps can be pre-specified or planned online.
Online planning is provided by the IHMC footstep planner\cite{griffin2019footstepplanner} by specifying a frame relative goal stance.
This planner can be setup to run online, enabling dynamic planning in the environment, or run once during authoring to pre-plan the footsteps.
A procedural "turn-walk-turn" planning mode is also available for flat ground scenarios.
To adjust the walking speed and reliability, the operator is also able to tune the nominal swing and transfer duration of the steps.

\begin{figure}
    \centering
        \includegraphics[width=0.9\columnwidth]{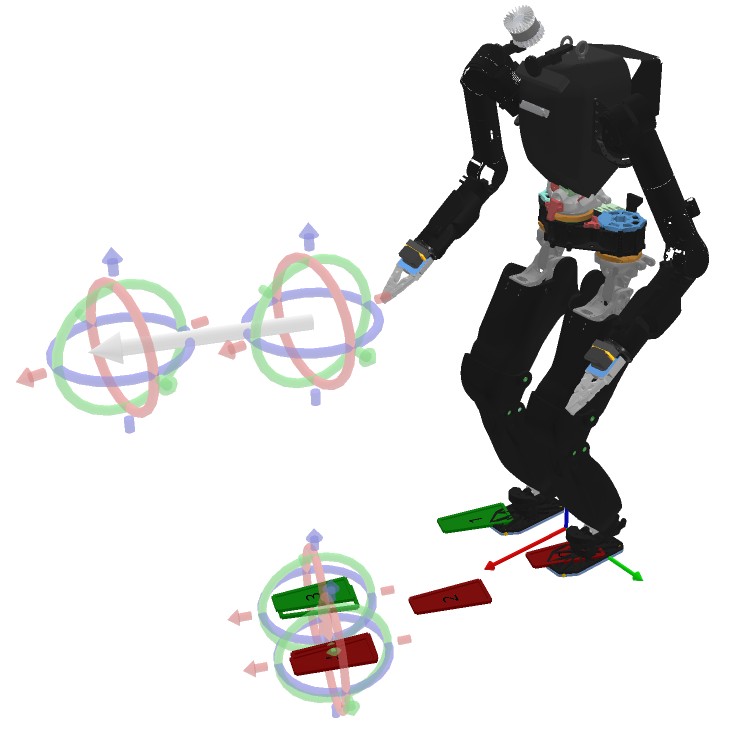}
    \caption{The approach stance AT is authored via a frame-relative point to stand under and point to specify the facing direction.
    The X-Y position and yaw of the stance footsteps are defined by the operator but are otherwise snapped to the ground.}
    \label{fig:ApproachStanceGizmos}
\end{figure}

We found that the perceived object frames can have significant enough orientation error that the goal stance steps do not align with the ground well enough for robust walking.
To account for this, when online planning is active, we snap the goal stance vertically onto the ground using two frame-relative points.
We define these as a point to face and a point to stand at.
These can be seen with a white arrow connecting them in \autoref{fig:ApproachStanceGizmos}.
The operator may also adjust the relative X-Y position and yaw of the stance steps with the gizmos as shown.

An arm trajectory action is used to move the arm and hand to either preset joint angles or an online IK solution for a frame-relative hand pose.
Specified for a single arm at a time (left or right), a 3D gizmo is used to adjust the desired hand pose in IK mode as shown in \autoref{fig:gizmo_composite}.
A trajectory time setting is used to adjust the speed of the motion.
To adjust the whole body controller's tracking priorities, tunable weights and gains are also provided.

\begin{figure}
    \centering
        \includegraphics[width=1.0\columnwidth]{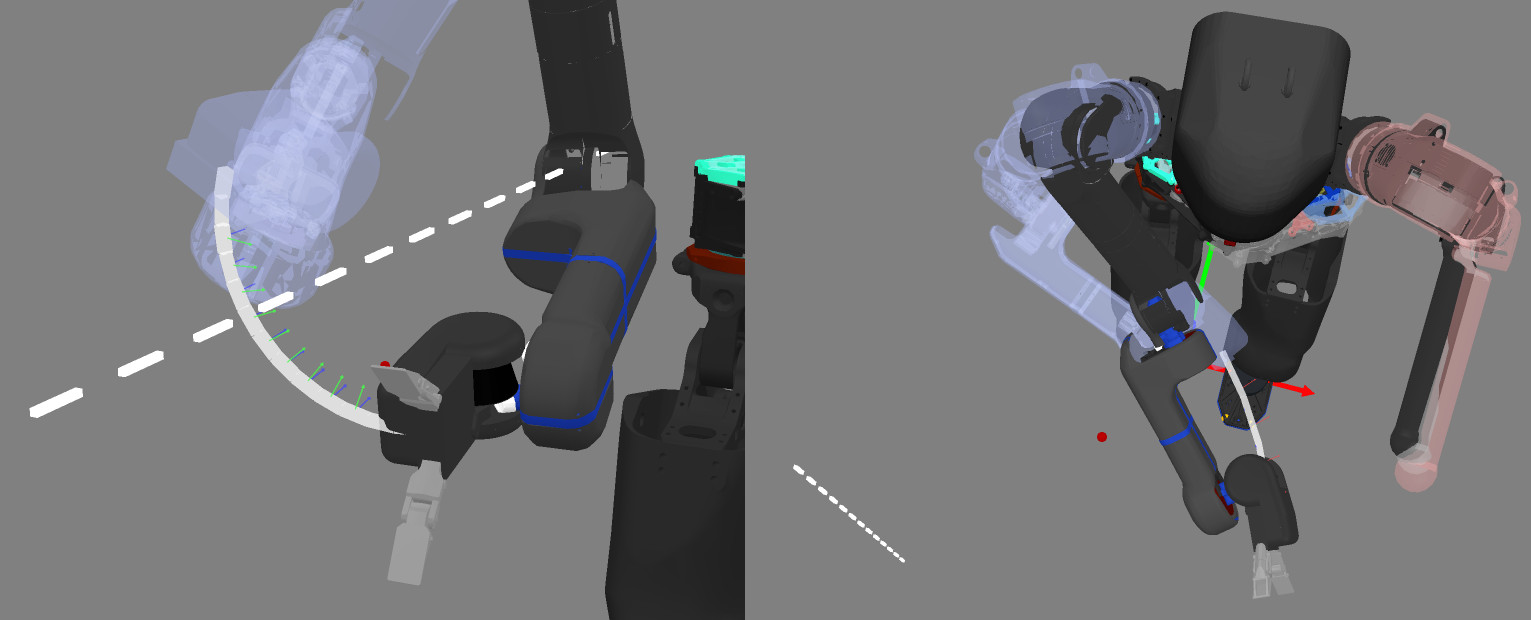}
    \caption{A visual rendering of generated screw primitive trajectories about two common axis configurations which are similar to those used for turning lever handles and opening doors by pulling the panel in a revolving motion.}
    \label{fig:ScrewPrimitiveTrajectoryGenerationUICombined}
\end{figure}

Inspired by \cite{Pettinger_2022}, we also implement a parameterized helical screw trajectory option, illustrated in \autoref{fig:ScrewPrimitiveTrajectoryGenerationUICombined}.
This allows the operator to define revolving motions useful for turning handles and swinging door panels.
It is defined by a 3D axis (dashed white line), an angle of revolution, and a translation amount to move along the axis.
The trajectory is also grasp-invariant, as it generates the trajectory online from the hand's pose just before execution.

Arm trajectories are often combined with chest and/or pelvis pose trajectory actions, which help to increase the reachability workspace.
For example, in the pull door behavior, the right arm moves across the robot to the handle on the left side of the door, aided by a chest yawing motion.

We control the hands with an action that tracks finger trajectories and preset hand configurations.
The operator can specify the joint angles of the knuckles or choose from a list of configurations such as "open" and "closed".
Additionally, a maximum torque option is available, which allows to operator to adjust the grip strength.

\subsection{Whole Body Control}

The behavior coordination layer was built to interface with a whole body and walking controller that is able to execute arbitrary trajectory requests and footsteps.
The behavior runtime process sends requests to the controller and monitors execution through statuses sent back from the controller.

\begin{figure*}
    \centering
        \includegraphics[width=0.8\textwidth]{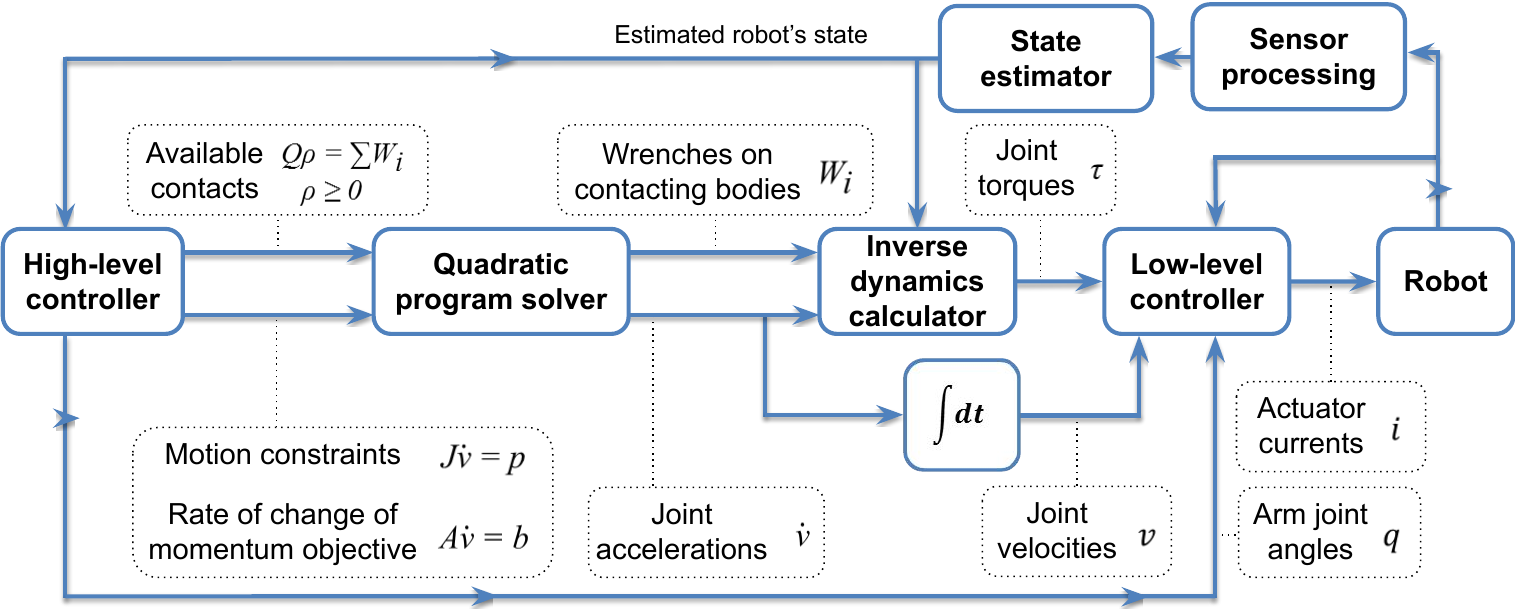}
    \caption{IHMC overall whole-body control framework.}
    \label{fig:ControllerOverview}
\end{figure*}

\begin{figure*}
    \centering

        \includegraphics[width=0.7\textwidth]{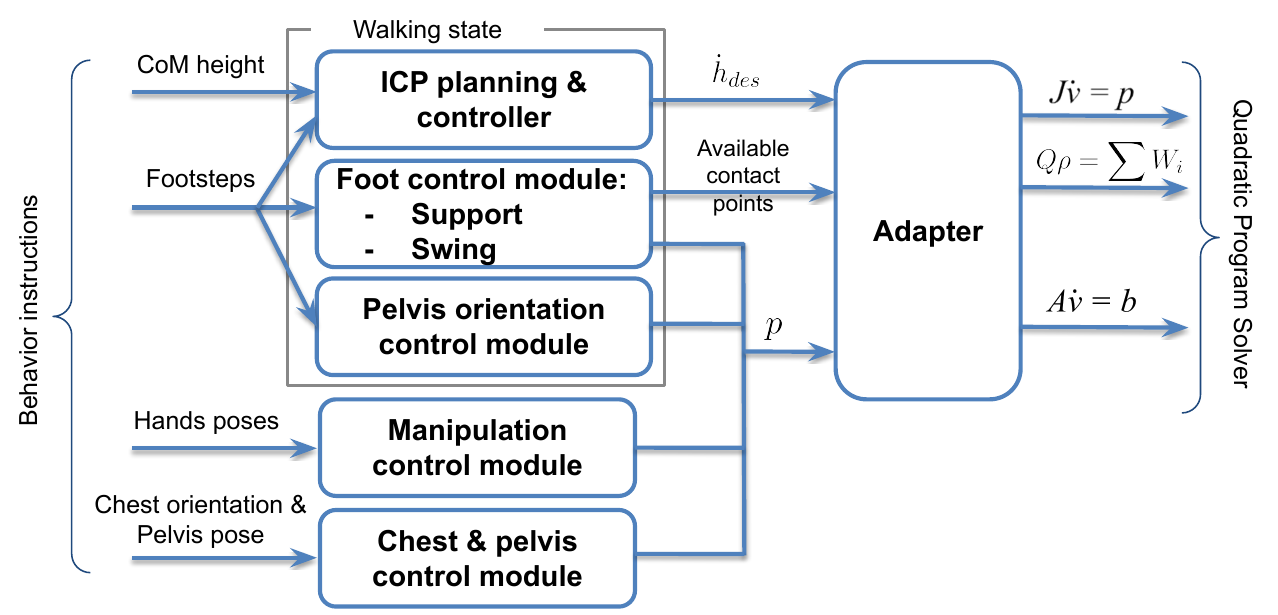}
    \caption{IHMC humanoid walking controller, highlighting several different control modules that regulate mobility and manipulation. Outputs of these control modules become input to the quadratic program solver in the whole-body control framework.}
    \label{fig:highLevelControllerOverview}
\end{figure*}

We use the whole body momentum-based control framework developed by Koolen et. al.\cite{Koolen_2016}.
The general structure and control flow of the whole body controller and walking implementation is illustrated in \autoref{fig:ControllerOverview} and \autoref{fig:highLevelControllerOverview}.
It is a model-based torque control scheme centered around a quadratic program (QP) that reconciles motion tasks expressed as constraints on the joint acceleration vector.
To manage the robot's interaction with the environment, we manage additional QP constraints for foot-ground contact and force-limited grasping with the hands.
The action primitives described previously make use of motion tasks provided by this controller which include jointspace and spatial (taskspace) acceleration objectives.

We use the dynamic walking behavior developed by IHMC and described in \cite{Koolen_2016}.
It is comprised of a state machine with standing, transfer, and swing states where each has associated motion tasks and active contacts that are used to achieve balance.
The controller is setup to execute a queue of footstep poses submitted by the user.

To model balance dynamics, it uses instantaneous capture point (ICP)\cite{pratt2006capturepoint}.
Foot pose, pelvis height, and pelvis orientation trajectories are generated to each achieve the user desired footsteps while maintaining balance.
During transfer an ICP trajectory is also generated, along with a relaxation of stance foot contact constraints to allow the heel to lift.
The centroidal moment pivot (CMP) is also be used to exhaust the body's ability to to control balance with its momentum.
More details can be found in \cite{Pratt2019_bookchapter}.

Foot swings are executed referencing smoothed trajectories which vary by terrain type.
On rough terrain, foot swings may be planned to avoid tripping on obstacles.
On flat ground, default, conservative trajectories are used that have low swing heights.
To recover from external disturbances and instabilities, swings can also be sped up during execution to more quickly place the foot and catch balance.

Lastly, for state estimation, a combination of a kinematic foot-ground contact proprioception for pelvis position and a pelvis-mounted IMU for pelvis orientation is used.
When a foot is in contact with the ground, determined by a glitch filtered and delay compensated ankle torque sensor based virtual foot switch, we assume the foot does not slide.
With one or both feet on the ground and assumed still, the pelvis position is computed simply via forward kinematics.
Consequently, the positional estimate of the robot is susceptible to drift in three main ways: inaccuracy of foot contact detection, foot slipping, and lack of both feet being on the ground.

\subsection{Operator Interface}

\begin{figure*}
    \centering
        \includegraphics[width=2.0\columnwidth]{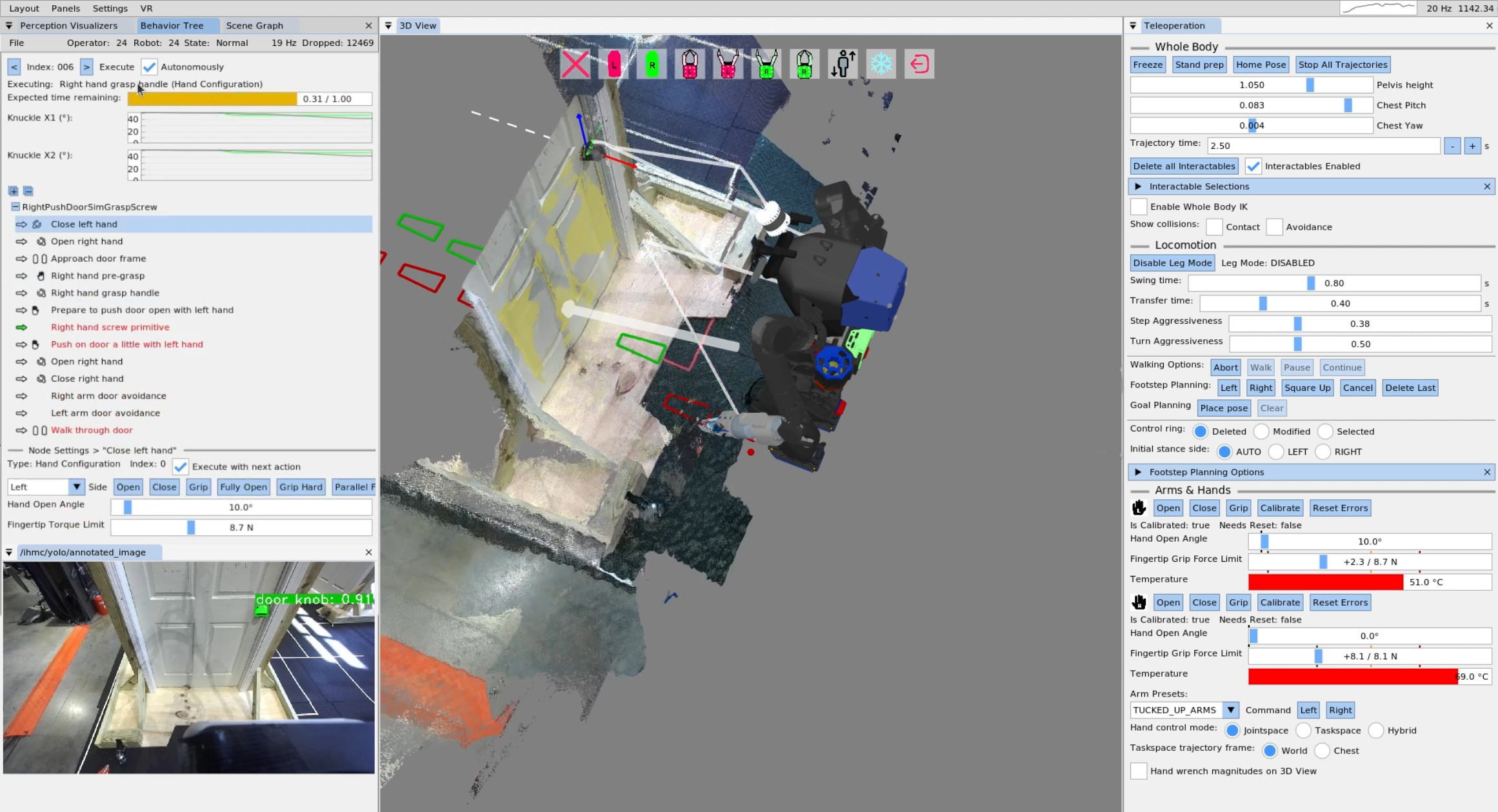}
    \caption{The operator interface used to author behaviors.
    On the left, the behavior tree editor allows to edit and operate the behavior tree.
    In the center, the 3D scene shows affordance template elements projected into a scene with the live robot state and perception data.
    On the right, the teleoperation panel provides controls to directly operate and monitor the robot.}
    \label{fig:20240701_NadiaBehaviorUI2}
\end{figure*}

Our operator interface is shown in \autoref{fig:20240701_NadiaBehaviorUI2}, which allows a human operator to construct and execute behaviors.
It contains a digital twin of the robot and the environment rendered in 3D as well as panels of interactable widgets for teleoperation and interacting with behaviors.
The operator can execute actions one at a time or autonomously.
While the robot is performing, tracking plots, log messages, flashing colors, and other items-of-importance are presented to the operator in real-time.

A unique element of our interface is that it allows behavior creation, modification, and execution online.
This feature speeds up the development process as there is no re-deploy step to run a new behavior.
This means that changes to any part of behaviors can be tried immediately and iterated on quickly on the real robot.
This ``on the fly" authoring also supports the operator-robot team to accomplish and automate never before seen tasks as they are encountered.

To preview what the robot will do next, we render a transparent robot and white trajectory graphics, which go through the motions of upcoming actions.
This helps the operator understand the resultant motions of primitive actions, the spatial occupancy of the robot, and the task-relative alignment of the goal pose.
Widgets are available to scrub through and playback the preview.

\begin{figure}[h]
    \centering
        \includegraphics[width=1.0\columnwidth]{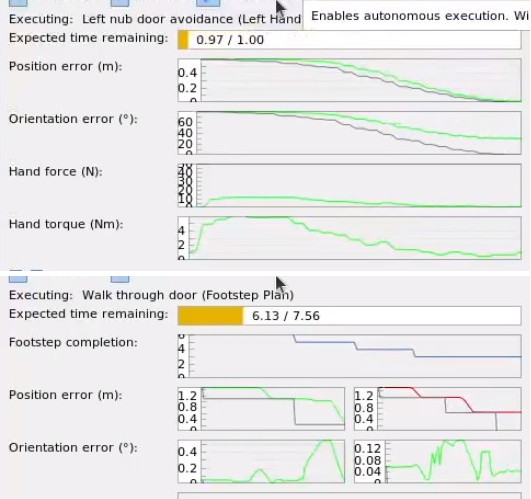}
    \caption{Live tracking plots in the UI help the operator monitor progress, identify failures, and design improvements.
    Top: Good tracking of a arm-hand trajectory.
    Bottom: Good tracking of a footstep plan. Discrete jumps represents steps.}
    \label{fig:TrackingPlots}
\end{figure}

As shown in \autoref{fig:TrackingPlots}, action execution tracking is plotted in real-time to keep the operator informed about progress.
We plot the Cartesian distance to the goal as one value rather than the separate components X, Y, and Z to help the operator focus on the overall tracking error.
A progress bar shows the remaining nominal execution time.
Red and green are used to indicate error and good tracking, respectively to further assist the operator.
A log area is also available that shows printed, colored messages from the robot's execution process.

Behaviors are stored as hierarchical JSON files which can be selected and loaded and saved.
Once the root node is created or loaded, the operator is presented with the UI shown in \autoref{fig:20240701_NadiaBehaviorUI2}.
Context menus allow the operator to add, remove, and reorganize nodes in the tree.
The tree and all of the nodes' properties and execution data is automatically and continuously synchronized with the robot's working copy.

There are two main modes for executing behaviors.
To test and iterate on behavior parts, a manual step-by-step mode is used by clicking a button to execute the next action.
The operator has an option whether to allow action concurrency in this mode.
The other main mode is the autonomous execution mode, where the robot will continue to execute actions as decided by the behavioral logic.

\section{Application to Door Traversals}

In this work, we consider door traversals where the door is closed at the beginning.
This is partly enabled by a perception system which can detect door opening mechanisms.
Door behaviors are generally composed of a sequence of approach, grasping and unlatching the door, swinging open the panel, and walking through.

\begin{figure*}
    \centering
        \includegraphics[width=2.0\columnwidth]{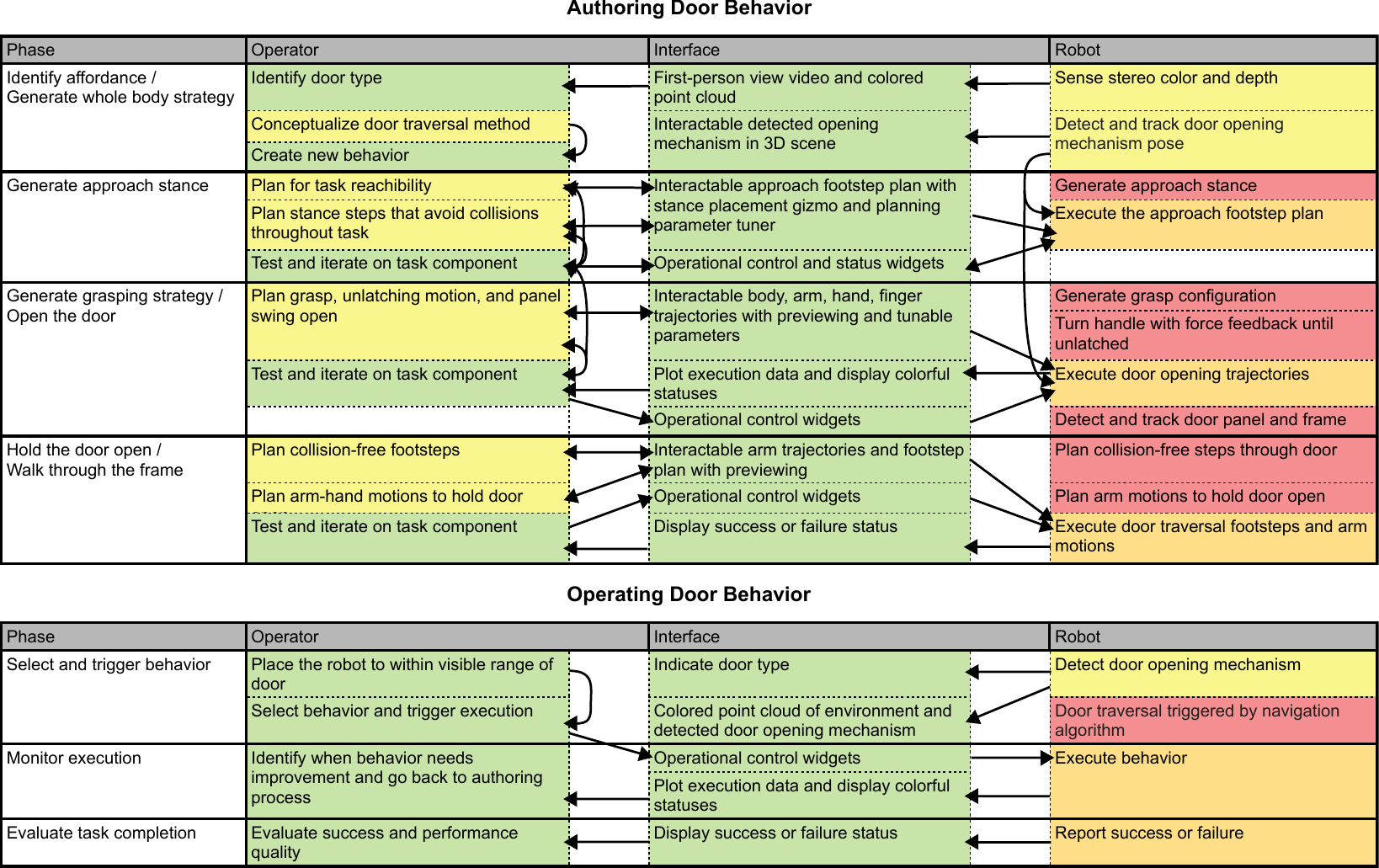}
    \caption{IA charts for authoring and operating pull door behaviors.}
    \label{fig:PullDoorBehaviorIACharts}
\end{figure*}

Many features of our behavior architecture have been developed through a Coactive Design (CD) process focusing on the human and robot team.
The Interdependence Analysis (IA) charts in \autoref{fig:PullDoorBehaviorIACharts} show connections between components involved in authoring and operating the door traversal behaviors.
The arrows pointing left towards the operator represent observability and predictability while the arrows pointing right toward the robot represent directability.
Red cells in the robot column represent capabilities that, if added, would improve reliability or efficiency.

\begin{figure}[h]
    \centering
        \includegraphics[width=1.0\columnwidth]{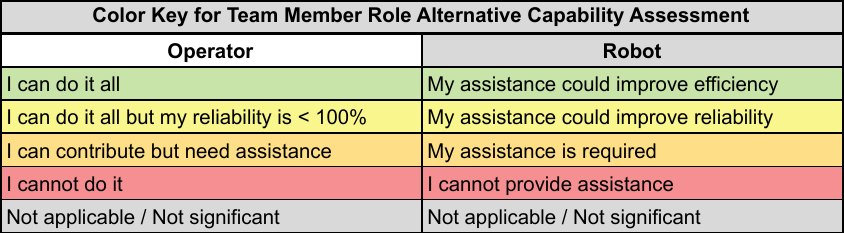}
    \caption{IA chart color key for team member role alternative capability assessment.}
    \label{fig:IAChartKey}
\end{figure}

The door approach task is defined by stance footsteps with respect to the pose of the door opening mechanism, which can be detected by the perception system or manually specified by the operator.
Online footstep planning is used to accommodate a wide range of initial robot stances.
For a push door, a squared up or staggered stance can be used, but for a pull door, the feet must be out of the way of the panel as it swings toward the robot.
A good general stance is off to the side of the door that the opening mechanism is on, with staggered feet.

The process of tuning the grasp of the door opening mechanisms is one of the most sensitive components, since many handle and hand combinations require a very precise alignment.
The operator typically uses an iterative process of manually tuning and triggering the grasp action until the result is reliable.
Grip force, trajectory duration, and hand orientation are primary parameters for this part.

Once the handle grasp action is ready, for a turnable knob or lever handle, the operator will use two screw primitive actions to unlatch and open the door.
To simplify the process, the screw primitives are designed to be grasp invariant.
Tuning the screw primitive actions primarily involves aligning the screw axes, shown as white dotted lines in \autoref{fig:ScrewPrimitiveTrajectoryGenerationUICombined}, to the axes of rotation of the handle and panel.
Once aligned, the angle of rotation is determined and the motions are iterated on until satisfactory.

The next and final phase of a door traversal is to walk through while using the arms to keep the panel out of the robot's way.
The first element of this is to persistently keep an arm in the way of the door closing again to prevent it from re-latching.
The second element is to walk through, with an arm or shoulder designated as a contact surface which keeps the door panel out of the robot's way.

\begin{figure}[h]
    \centering
        \includegraphics[width=1.0\columnwidth]{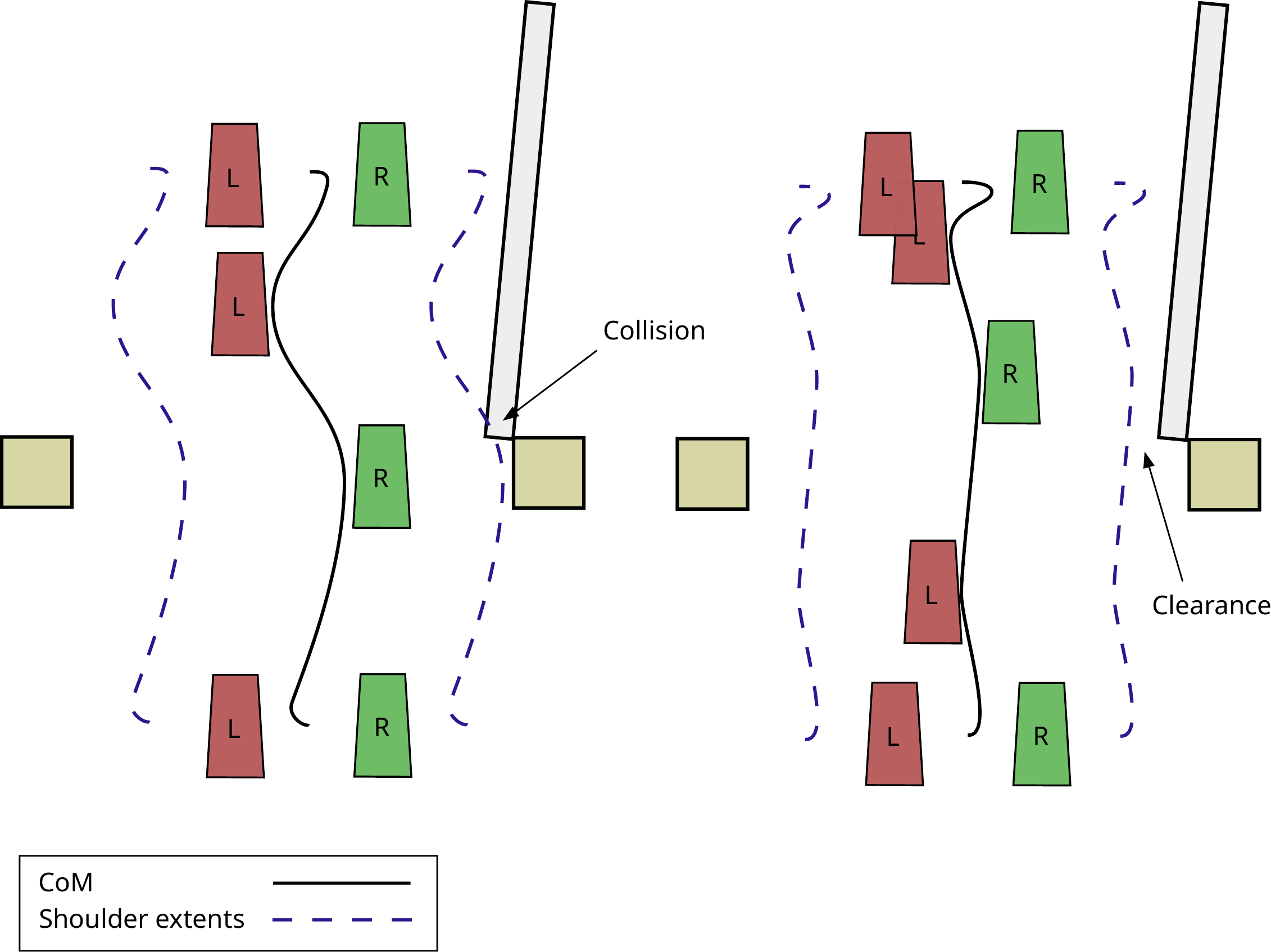}
    \caption{When walking through the door, narrower steps result in the robot's body (CoM) swaying less and resulting in collisions with the door frame.
    Additionally, straddling the door threshold with the the footsteps keeps the body centered in the door frame during the transition.}
    \label{fig:ShouldersDoorFrameFootsteps}
\end{figure}

Our robot's shoulders were 85\% as wide as the door frame opening which meant our normal walking footstep width resulted in collisions between the robot's shoulders and the door frame, causing the robot to fall.
As illustrated and modelled in \cite{2015_englsberger_dcm}, the center of mass (CoM) of the robot, located near the torso, will sway side to side during walking as an artifact of balancing.
As shown in \autoref{fig:ShouldersDoorFrameFootsteps}, both taking narrower steps and having the steps straddle the door threshold help to keep the shoulders centered with respect to the door frame and avoids collisions.
To do this, we utilized the manual step placement feature of the footstep plan action primitive.

\section{Door Handle Detection}

\begin{figure}[h]
    \centering
        \includegraphics[width=1.0\columnwidth]{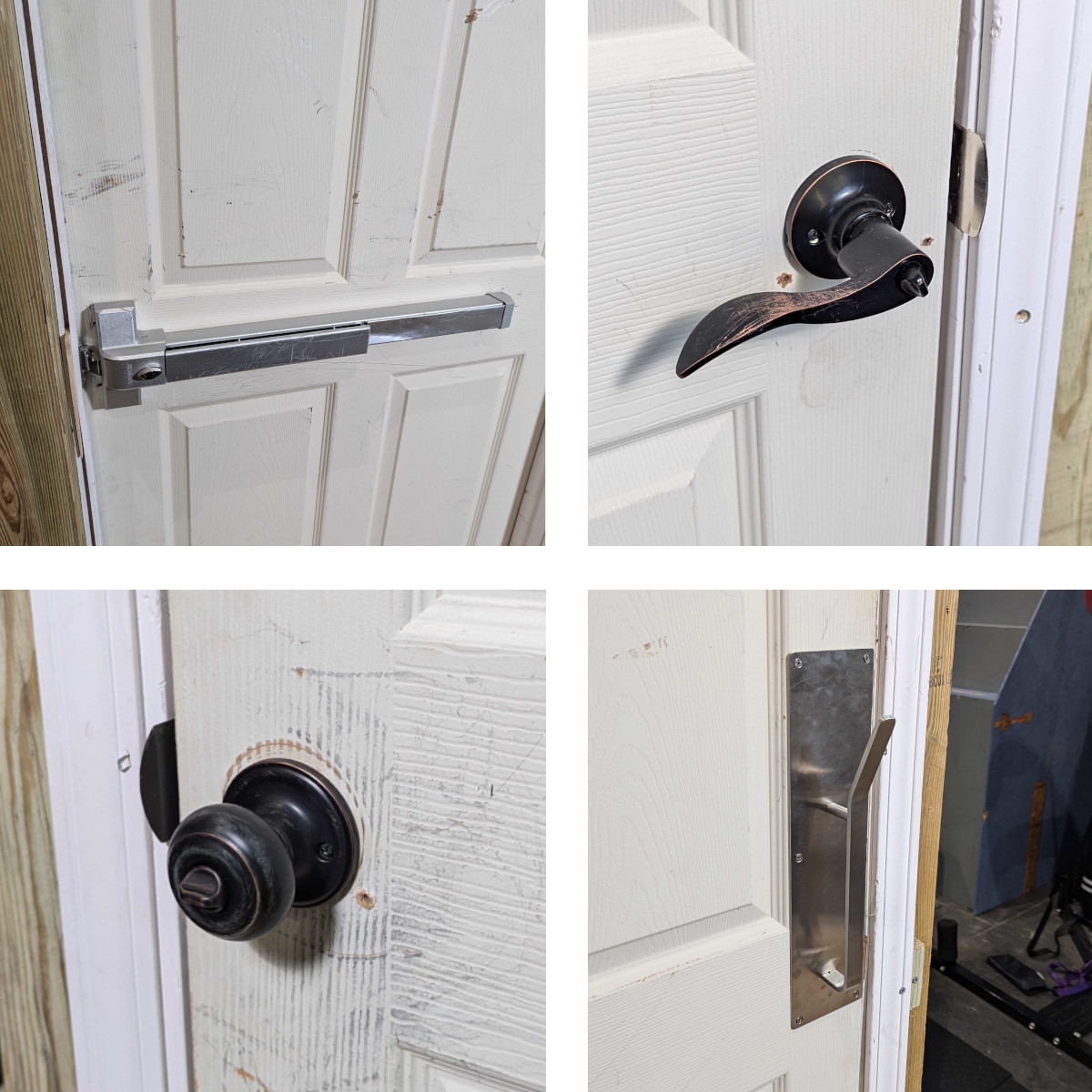}
    \caption{The four types of door opening mechanisms used: a push bar (top left), a lever handle (top right), a knob (bottom left), and a hands-free pull handle (bottom right).}
    \label{fig:DoorHandleTypes}
\end{figure}

To enable this presented behaviors, we developed a novel detection and tracking system for door opening mechanisms.
When tracked, the poses of detected door opening mechanisms are provided to the behavior runtime for use in authoring and execution of behaviors.
We developed models to detect and track the four different opening mechanisms shown in \autoref{fig:DoorHandleTypes}.

\begin{figure}
    \centering
        \includegraphics[width=1.0\columnwidth]{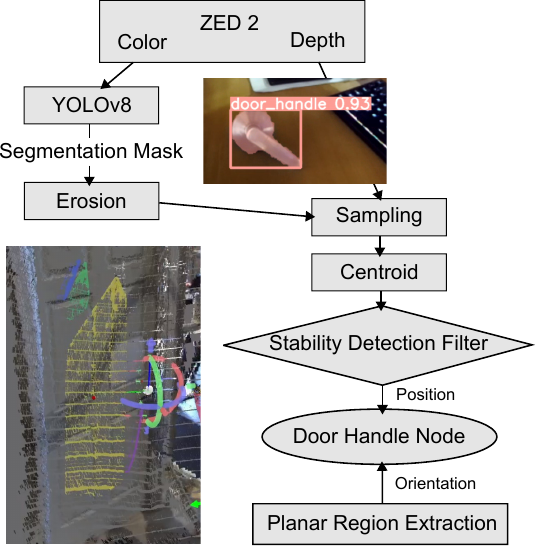}
    \caption{Our detection process uses ZED 2 color and depth data, YOLO, centroid calculation, stable detection filtering, and fusion with planar regions to estimate the pose of the door mechanism.}
    \label{fig:YoloManagerLayout}
\end{figure}

To detect and track the pose of door opening mechanisms, we use an 8 step process illustrated in \autoref{fig:YoloManagerLayout}.
We use a ZED 2i\cite{zed2} facing forward and downward at $43^\circ$ to readily view the area in which door manipulation occurs.
The color data from the sensor is fed through YOLO to obtain a segmentation of the mechanism and a planar region extractor is used to obtain the plane of the door panel.
The centroid of the depth points on the mechanism is used for the position and the normal of the plane of the door panel is used for the orientation.

\begin{figure}
    \centering
        \includegraphics[width=1.0\columnwidth]{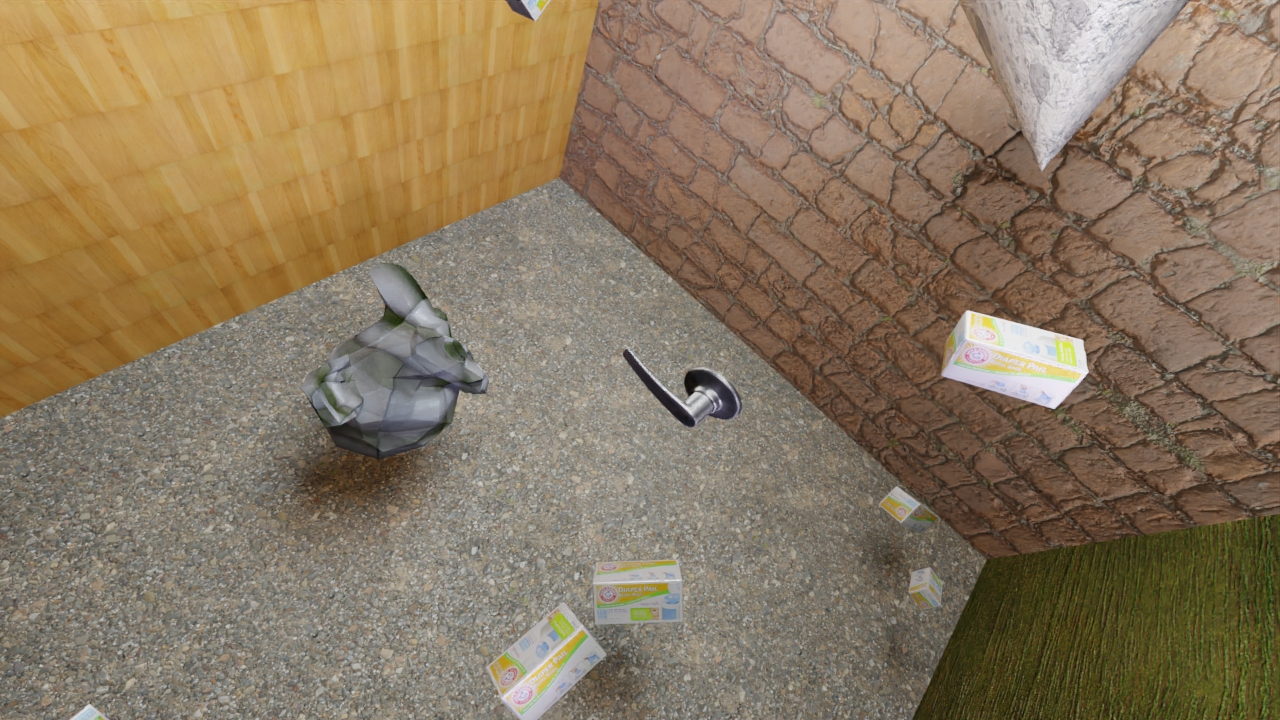}
    \caption{An example rendering of a door lever handle in Blender with randomized background and distractor placement.}
    \label{fig:DoorHandleSyntheticRender}
\end{figure}

In this work we re-trained YOLOv8\cite{yolov8} with data for our specific handle types.
The generate synthetic data for training, we scanned objects using Polycam\cite{Polycam} and rendered them in photo-realistic scenes using Blender\cite{blenderproc2}.
In each scene we also placed objects to serve as distractors, including primitive shapes and selections from Google Scanned Models.
An example render is shown in \autoref{fig:DoorHandleSyntheticRender}.
Thousands of images are rendered and associated with the corresponding ground truth object pose, binary segmentation image, occlusion-aware bounding box, and camera intrinsics.

To get a high quality result, we also included annotated real world data.
This data was gathered with the ZED 2 by capturing images of the door and handles with varied door positions, openness, lighting conditions, and background.
We obtained the segmentation labels by manually selecting the region of interest and running Segment Anything\cite{kirillov2023segment} on that area.
To improve training, for 15\% of the samples, we used images that did not contain a door or a handle.
We found that a combination of 1000 real and 1000 synthetic labeled samples were enough to train YOLO for sufficient performance.

To calculate the position of the door opening mechanisms, the resulting YOLO segmentation is eroded and combined with the corresponding depth points from the sensor.
The points are sampled down to a smaller number and a centroid is taken.
The centroid is then alpha-filtered to reduce noise and improved the stability of the result.

To obtain the orientation of the mechanism, the rapid planar region extractor from \cite{Mishra_2021} is run with the depth data from the ZED 2i sensor.
For a mechanism detected by YOLO, we take the positional calculation and search for a planar region that is nearby, has a normal perpendicular to gravity, and is sufficiently large to plausibly be the door panel.
If such a planar region is found, the orientation is set as the mechanism's pose.
To provide the behaviors with a stable and reliable detection, we require that both YOLO detects the mechanism and a valid planar region association is made for at least 60\% of the sensor frames within a one second window.

\section{Results}


The videos demonstrating our results are available \href{https://www.youtube.com/playlist?list=PLXuyT8w3JVgMPaB5nWNRNHtqzRK8i68dy}{here}.
We performed six variations of door traversals by varying push and pull with left and right handed hinges and with knob, push bar, and lever handle opening mechanisms.

We performed these experiments on the Nadia humanoid robot developed by IHMC with Boardwalk Robotics.
It features 6 degree of freedom (DoF) legs, a 3 DoF spine, 7 DoF arms, and 1 DoF grippers.
The ankle, knee, hip pitch and roll, and spine pitch and roll joints are actuated by Moog Integrated Smart Actuators (ISAs)\cite{Moog2023}.
The hip and spine yaw joints are actuated by electric motors with planetary drives.
The arms are powered by cycloidal drive actuators, described in \cite{bertrand2024fastteleop}, which provide high torques, near-zero backlash, and robustness to heavy impacts.
SAKE EZ Grippers\cite{sakerobotics} are used for the hands.
The robot features a carbon fiber exoskeleton for the thighs, shins, and torso to save on weight and enable contact with the environment.

Because our robot is tethered to a hydraulic pump and cannot sustain falls, we built door mock ups without the top part of the frames.
The lab doors are otherwise designed using real door hardware commonly found in residential and commercial buildings.
Additionally, spring closers are mounted and were used for some of the experiments.

\begin{figure}
    \centering
        \includegraphics[width=0.5\columnwidth]{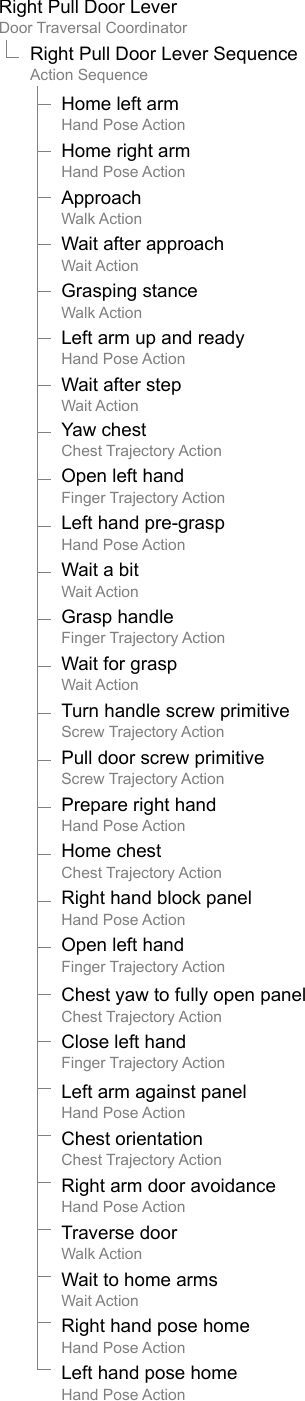}
    \caption{The tree for the right pull lever door traversal behavior. Note the actions are listed sequentially but scheduled dynamically and concurrently as shown in \autoref{fig:ActionLayeringIllustration}.}
    \label{fig:RightPullDoorLeverTree}
\end{figure}

Video key frames of a right pull lever door traversal are shown in \autoref{fig:2024NadiaRightPullLeverHandleCycloid}.
The corresponding tree is illustrated in \autoref{fig:RightPullDoorLeverTree}.
The sequence is composed of three main phases: approach, unlatch and open door, and walk through.
Wait actions are added to allow the perception system to get a more accurate pose of the door handle and increase reliability.

\begin{figure}
    \centering
        \includegraphics[width=1.0\columnwidth]{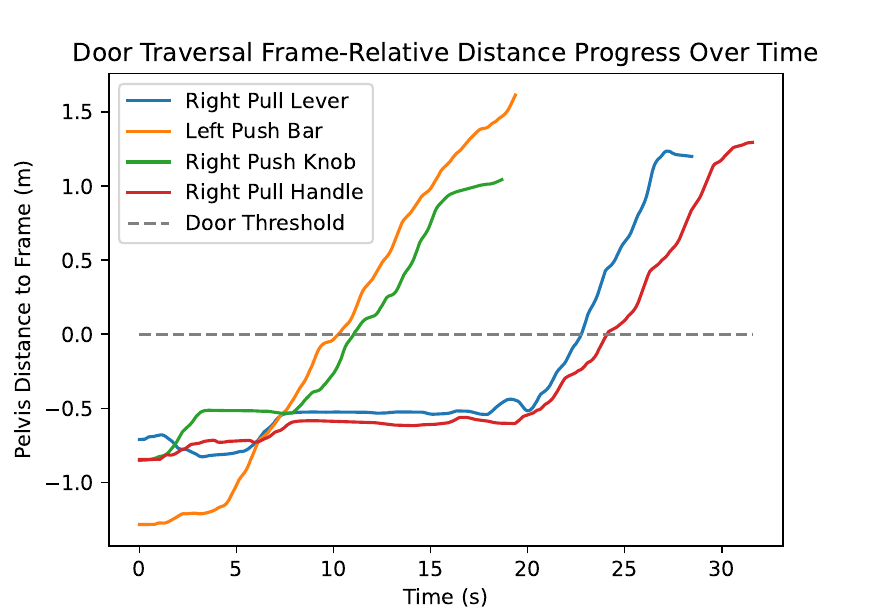}
    \caption{Plots showing behavior speed as distance progress through the door frame. Four door behavior runs on real robot of various types are shown.}
    \label{fig:DoorTraversalTimings}
\end{figure}

Our door traversal behaviors execute on the timescale of tens of seconds.
The distance over time of four representative types of door traversal behaviors on the the real robot are plotted in \autoref{fig:DoorTraversalTimings}.
Push door behaviors are the quickest, executing in less than 20 seconds, because the robot can keep making forward progress.
Pull door behaviors take about 10 seconds longer because the robot must stop further back, stay clear of the door panel, and open it fully before proceeding.

\begin{figure*}
    \centering
        \includegraphics[width=1.8\columnwidth]{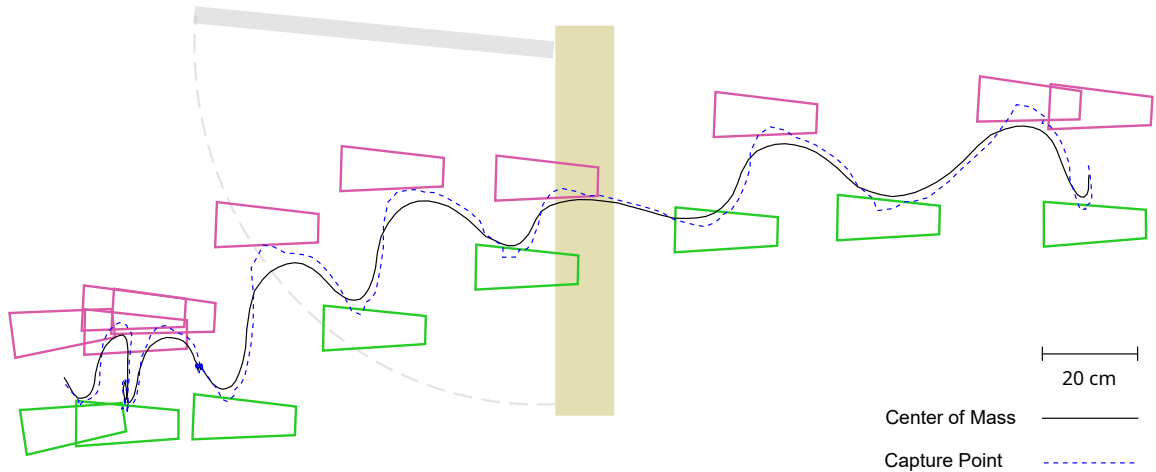}
    \caption{An overhead plot of a right pull handle door traversal. The footsteps and center of mass and instantaneous capture point trajectories are plotted to show balance performance.}
    \label{fig:RightPullHandleOverheadPlot}
\end{figure*}

During traversals with spring loaded door closers, the robot can experience significant disturbances.
To show the degree of balance disturbance, an overhead plot of the center of mass and instantaneous capture point trajectories for a right pull handle door traversal is shown in \autoref{fig:RightPullHandleOverheadPlot}.

\begin{figure*}
    \centering
        \includegraphics[width=2.0\columnwidth]{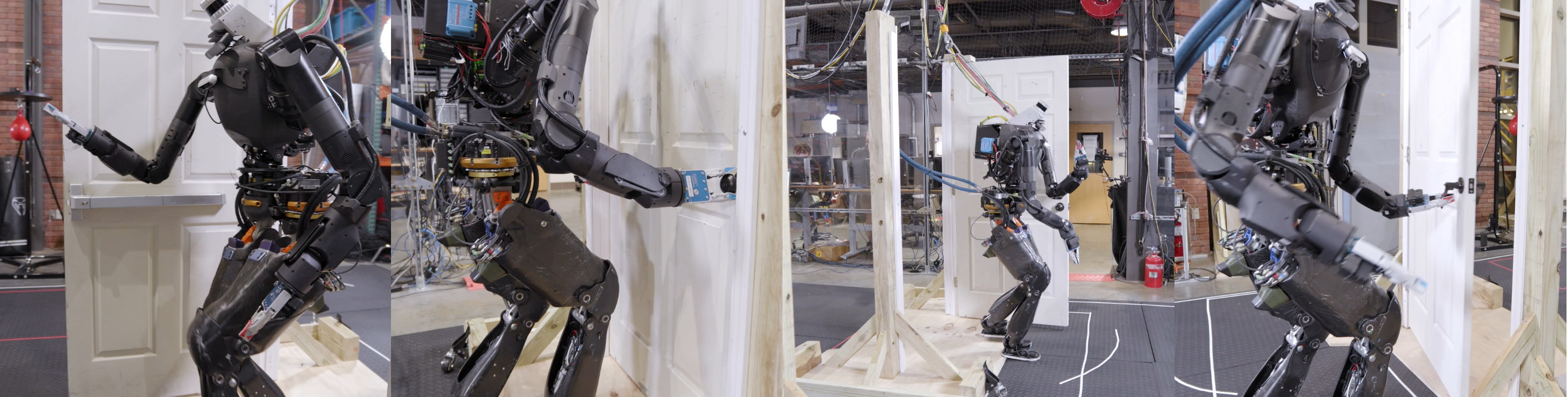}
    \caption{Three door traversals in a row with continuous autonomy.
This illustrates the composability of higher level building exploration behaviors.
The door traversal types are push bar door right hinge side, right push knob, and right pull lever handle.
    Corresponding video: 202407NadiaDoorTraversalsPushBarPushKnobPullHandleFallLowRes.mp4}
    \label{fig:202407_NadiaDoorTraversalsPushBarPushKnobPullHandleFall}
\end{figure*}

We also ran three door traversals consecutively in a single composite autonomous behavior.
Video key frames of this three-in-a-row door traversal run are shown in \autoref{fig:202407_NadiaDoorTraversalsPushBarPushKnobPullHandleFall}.

\begin{figure}
    \centering
        \includegraphics[width=0.8\columnwidth]{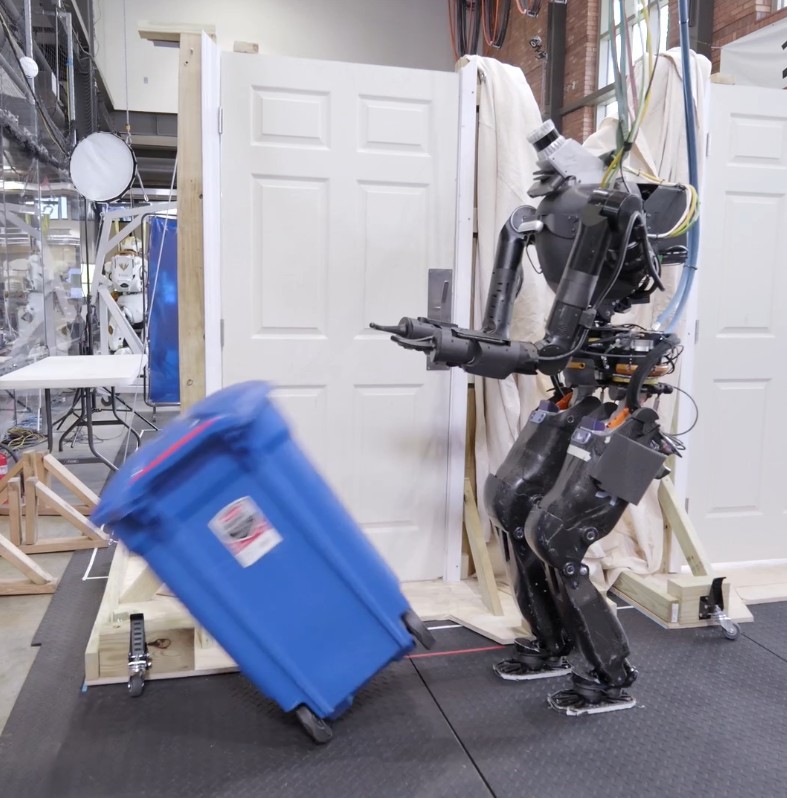}
    \caption{A behavior that moves a recycling bin that is blocking access to the doorway.
    Corresponding video: 202407NadiaClearTrashCanLowRes.mp4}
    \label{fig:2024_ClearTrashCan}
\end{figure}

To demonstrate the generality of the approach, we included a debris clearing behavior as part of a right pull handle door traversal.
In \autoref{fig:2024_ClearTrashCan}, the robot can be seen executing this behavior which pushes a recycling bin out the the way of the door.
This experiment included a building exploration node that identified that the recycling bin was blocking the door, executed the behavior to move it out of the way, and the executed the door traversal behavior.

\section{Discussion and Future Work}

Our work has enabled a bipedal humanoid robot to traverse doors in 20-30 seconds, which is many times faster than the prior methods used by the IHMC Robotics lab, including in the DARPA Robotics Challenge Trials in which IHMC Robotics won the ``Best in Door'' award.
We think a variety of factors lead to this dramatic increase in speed.
First, the Nadia robot hardware features higher maximum joint velocities than DRC Atlas and NASA Valkyrie, enabling it to walk and move faster.
Second, the 7 DoF electric arms with cycloidal drives provide good compliance, high torques, and high speeds, enabling manipulation motions to be fast and robust to unplanned contacts.
Third, the concurrent action implementation discussed in this paper enables the robot to execute layered actions as soon as dependent actions complete, eliminating pauses and doubling speed.
Fourth, the coactive UI design has allowed for rapid iteration of these behaviors, enabling us to find and implement faster strategies more quickly.
Last, advances in GPU accelerated perception have allowed the robot to identify door features both directly and quickly, removing even more long pauses.

Despite great results in fast door traversals and a framework for generating general loco-manipulation behaviors, there is still a considerable amount of improvements that could be made.
These improvements focus mainly on increasing executing speed, decreasing dependence on a human operator, and increasing the capability to perform both dexterous and high powered manipulations.

We are still far from achieving human speed and performance, which would be necessary for a robot to stay with a group of humans in urban operations.
Major slowdowns in this work is unnecessary footsteps, pauses, and maximum walking speed.
One way to mitigate this would be to eliminate steps that are not necessary and keep the robot walking by blending the footstep plans between actions.
The robot should only take steps that are purposeful and of utility.

In this work, we defined trajectories primarily in position space, but we believe moving to force and impact based action definitions will simplify the action definition, increase disturbance robustness, and provide another way to perceive the environment, through proprioception, by gathering reaction force data.
Doors, being constrained by revolute joints, would be a fitting candidate for such techniques.
Work has been done in this area in the form of impedance controlled actions \cite{Jain2010EPC}\cite{Pettinger_2022}.

We also relied heavily on the ingenuity of a human author and operator.
One way to mitigate that dependency is to implement more advanced planners, such as stance and grasp generation and collision free motion planning as in \cite{2010ChittaDoorOpening}.
\cite{2023JangDoorTraversal} uses a graph based planner combined with an IK trajectory solver for navigating door traversals for wheeled base robots with an arm.
\cite{zhang2024learningopentraversedoors, kang2024door} use reinforcement learning to train a quadrupedal robot with an arm to traverse doors.

Lastly, we are unable to detect the door frame and thus must assume the door is closed when it's first seen.
A detection algorithm for door frames would enable traversing doors in any initial condition.
We speculate that some additional planning heuristics would also be required for doors that are initially ajar.

\section{Conclusion}

We have presented theory, designs, an implementation, and real world results for a humanoid robot loco-manipulation behavior architecture.
A key defining characteristic of our work is the speed of the robot in accomplishing tasks.
We think our system is the fastest presented such system in the literature.
Ultimately, achieving human speeds and faster in building exploration and other loco-manipulation tasks are key for humanoid robots to become useful.

\subsection{Acknowledgements}

We would like to thank Jerry Pratt, Stephen Hart, Matt Johnson, William Howell, Thomas McKenna, and the IHMC Robotics team, without whom this work would not have been possible.

\subsection{Source Code and Media}

Our implementation and the associated modules discussed in this paper can be found on our GitHub at \url{https://github.com/ihmcrobotics}.
The accompanying videos can be found \href{https://www.youtube.com/playlist?list=PLXuyT8w3JVgMPaB5nWNRNHtqzRK8i68dy}{here}.

\section{Bibliography}

\bibliographystyle{cas-model2-names}

\bibliography{cas-refs}

\vskip50pt

\bio{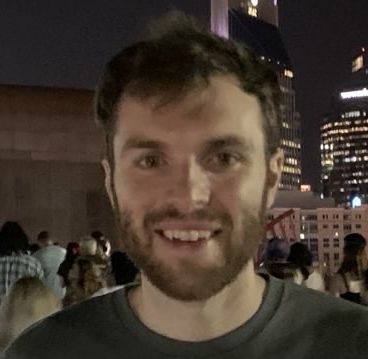}
\textbf{Duncan Calvert} received a B.S. in Computer Science at the University of West Florida (UWF) in 2014.
He has worked as a Research Associate in the Robotics Lab at IHMC since 2014.
He served as the operator's copilot in the DARPA Robotics Challenge Finals in 2015, where IHMC took 2nd place.
He is now a PhD student in the Department of Intelligent Systems and Robotics (ISR) at UWF.
His research interests include legged locomotion, manipulation, behaviors, and teleoperation for humanoid robots.
\endbio

\bio{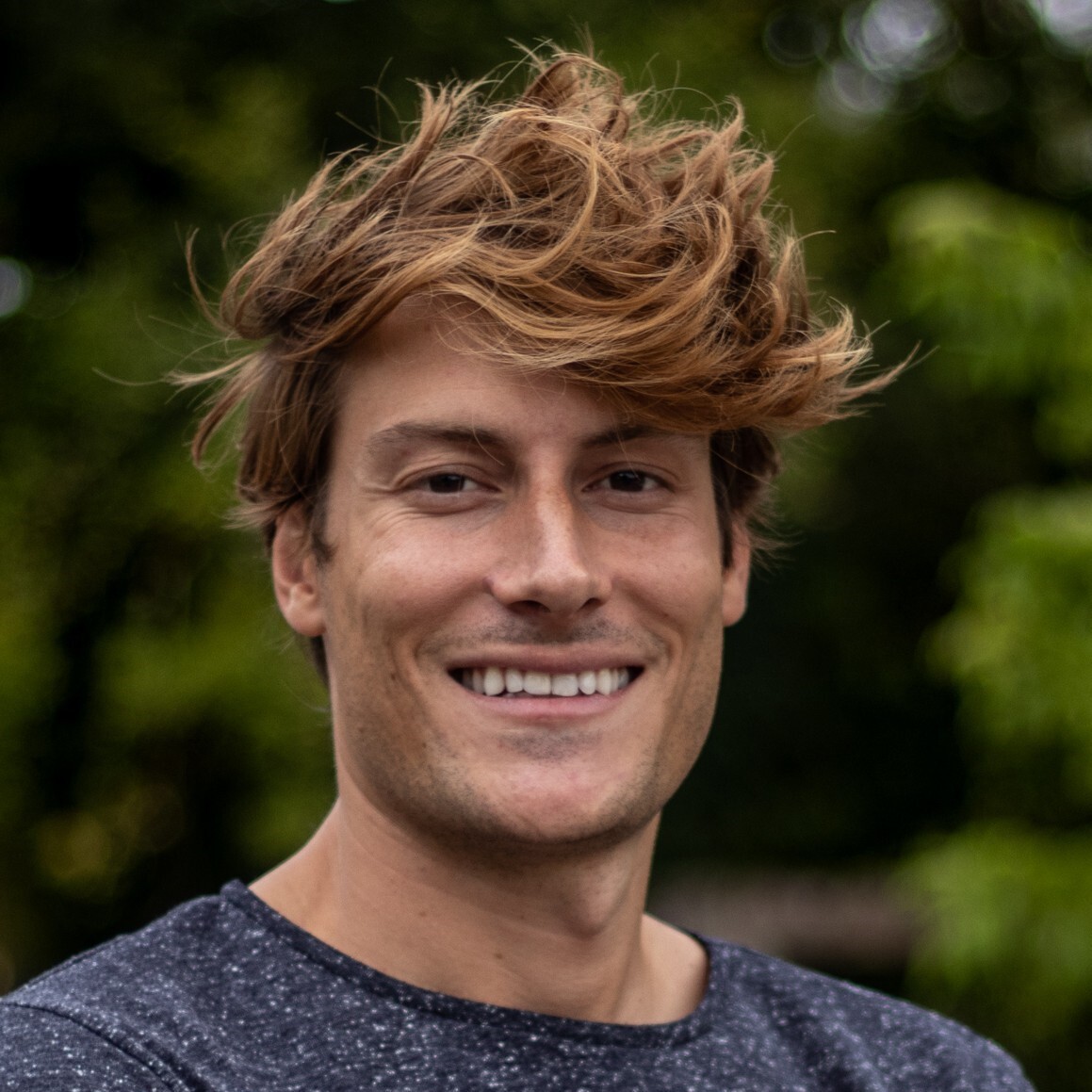}
\textbf{Luigi Penco} is a reserach scientist at IHMC and focuses on developing autonomous and semi-autonomous behaviors for humanoid robots, as well as user interfaces for their control.
He earned a bachelor’s degree in Electronics Engineering from Roma Tre University in 2015 and a master’s in Artificial Intelligence and Robotics from La Sapienza University of Rome in 2018.
In 2022 he received a PhD in robotics from Université de Lorraine, while conducting his doctoral studies at Inria Nancy Grand-Est.
His research interests include teleoperation systems, impact-aware control strategies and behavior generation, coordination and adaptation methods.
\endbio

\bio{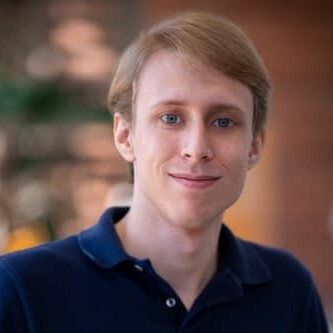}
\textbf{Dexton Anderson} is a Software and DevOps Engineer working with the humanoid robotics team at IHMC.
He is from Pensacola and received his Bsc. in Computer Science from UWF.
He contributes to various IHMC vendored software libraries used by the humanoid robotics platforms IHMC has access to and maintains continuous integration systems used by the robot lab.
Dex has previously worked for The National Flight Academy (NAS Pensacola) developing flight simulations and add-ons for Prepar3D flight simulator.
He also has created and contributes to many open-source software projects in his free time. 
\endbio

\bio{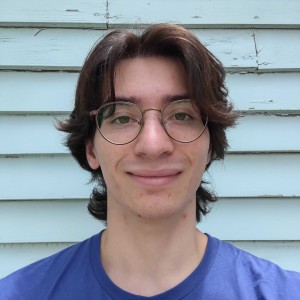}
\textbf{Tomasz Bialek} received his B.S. in Computer Science at the University of West Florida in 2024.
He has been an intern at IHMC since 2023.
He has worked on humanoid robot hand low level control and user interfaces, GPU acceleration of sensor data acquisition, and heuristic object detection algorithms that combine the results of multiple state of the art approaches.
He enjoys photography and working on perception algorithms.  
\endbio

\bio{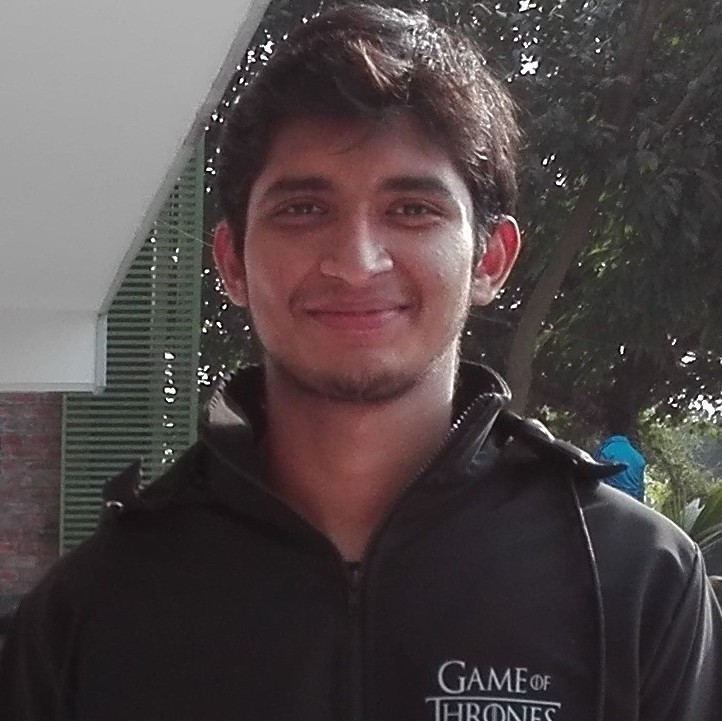}
\textbf{Arghya Chatterjee} received a B.S. in Mechanical Engineering (Robotics Major) at the Bangladesh University of Engineering and Technology (BUET) in 2019.
He has worked as a software engineer in Lotus Robotics where he developed autonomous UGVs for industrial security inspection and food delivery since 2020.
He was a perception team member of Team CoSTAR from NASA JPL during DARPA SubT Challenge where he worked on degeneracy-aware SLAM, Object Detection and Localization techniques for Single and Multi-robot autonomous systems in 2020-2021. 
He is now a PhD student in the Department of Intelligent Systems and Robotics (ISR) at UWF.
His research interests include object detection, pose estimation, object surface reconstruction and autonomous manipulation for humanoid robots.       
\endbio

\bio{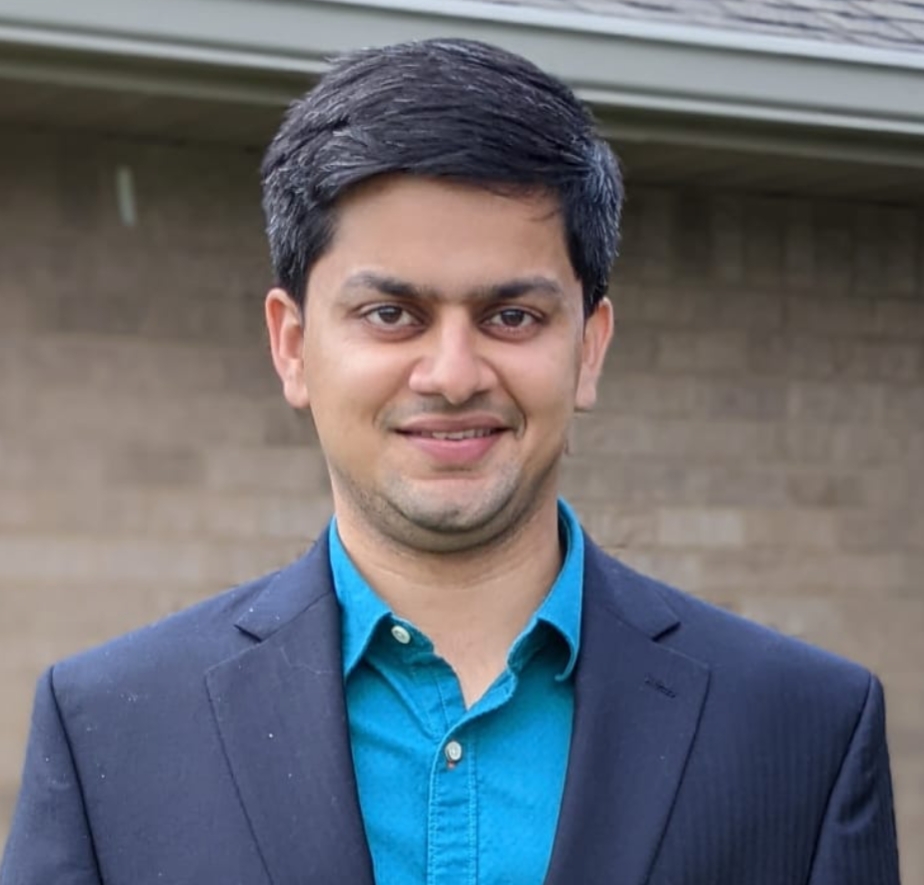}
\textbf{Bhavyansh Mishra} received a B.S. in Computer Engineering at the University of West Florida (UWF) in 2019.
During his undergraduate degree, he also published research on DevOps, distributed software architectures, and secure software development practices.
He has also received numerous awards in computer vision, cybersecurity, and autonomous drone racing competitions.
He also defended his PhD dissertation on perceptive bipedal locomotion in the Department of Intelligent Systems and Robotics (ISR) in 2024.
His research focused on perception and planning algorithms that enabled bipedal humanoid robots to traverse challenging terrain in real-world applications.
\endbio

\bio{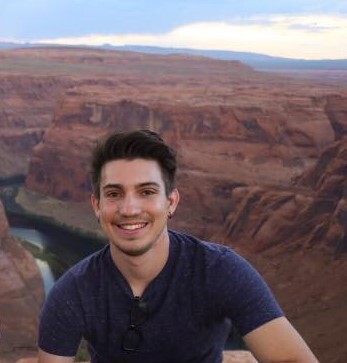}
\textbf{Geoffrey Clark} is a Research scientist at IHMC and focuses on machine learning and artificial intelligence research for humanoid robots.
In 2023 he received a PhD in Electrical Engineering from Arizona State University, while pursuing research in the Human Robot Interaction Lab.
During his doctoral research, Dr. Clark created generative probabilistic models of legged walking systems, including autonomous robots, prosthetics, and exoskeletons.
Using these models he examined how to utilize the probabilistic framework to optimize human and robot interactions for beneficial outcomes, such as reducing joint loads of amputees to reduce the risk of osteoarthritis.
\endbio

\bio{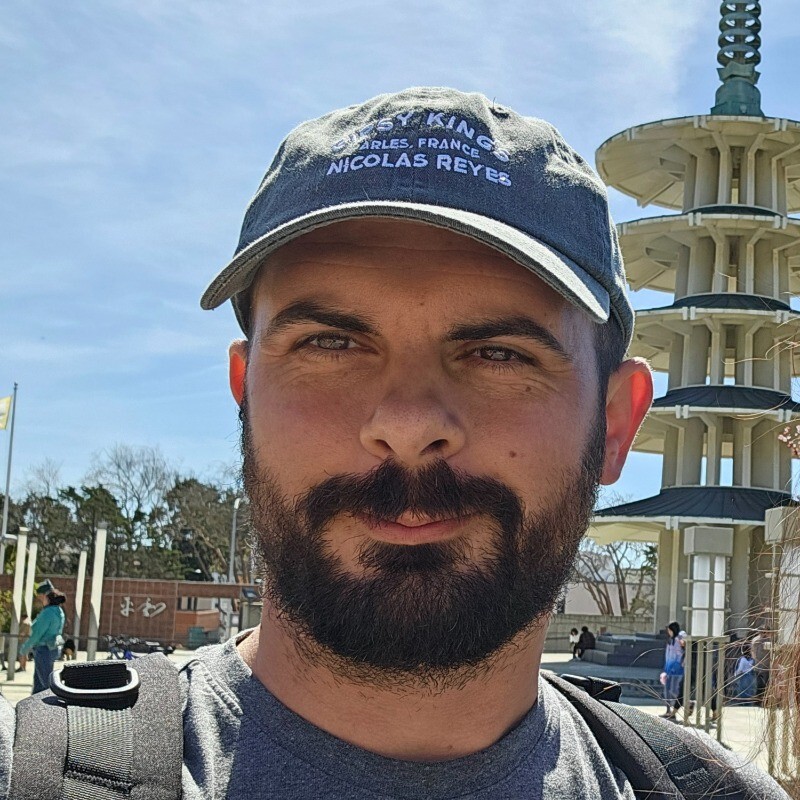}
\textbf{Sylvain Bertrand} received a PhD in Mechanical Engineering at the University of Versailles in 2013.
He has worked as Research Scientist at the Robotics Lab at IHMC since 2013.
He was the control lead during the DARPA Robotics Challenge from 2013 to 2015, where IHMC took 2nd place in the finals.
He has since contributed in several domains such as: whole-body controls, perception, manipulation, and teleoperation for humanoid robots.
\endbio

\bio{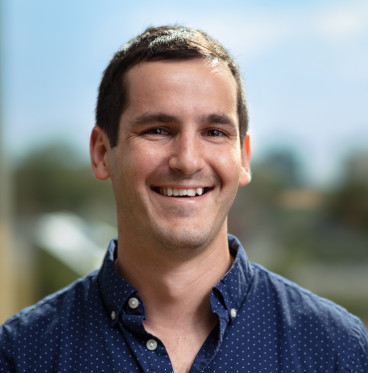}
\textbf{Robert Griffin} is a Research Scientist with IHMC, where he leads the Robotics team.
His research focuses on improving mobility and autonomy for legged robotics and powered exoskeletons.
He is interested in system level approaches for improving the mobility and capability of these robotic platforms, including platform design, motion design and control, autonomy and manipulation, and perception.
His work involves leading a number of projects, including developing the next generation humanoid robot, Nadia, which has a high power-to-weight ratio and large range of motion and will enable exploration and interaction with complex, dynamic, and unknown environments such as ships and urban settings.
Robert received his PhD from Virginia Tech is 2017.
\endbio

\end{document}